\documentclass{article}
\usepackage[T1]{fontenc}
\usepackage[utf8]{inputenc}
\usepackage{arxiv}
\usepackage{amsmath}
\usepackage{amssymb}
\usepackage{mathtools}
\usepackage{graphicx}
\usepackage{booktabs}
\usepackage{multirow}
\usepackage{longtable}
\usepackage{array}
\usepackage{adjustbox}
\usepackage{xcolor}
\usepackage[normalem]{ulem}
\usepackage{enumitem}
\usepackage{setspace}
\usepackage{subfig}
\usepackage{natbib}
\bibpunct[, ]{(}{)}{,}{a}{}{,}
\usepackage[hidelinks]{hyperref}
\newcommand{\sname}{OpenHail}
\renewcommand{\shorttitle}{OpenHail: Electric Ride-Hailing Fleet Control}
\title{OpenHail: An Event-Driven Gymnasium Environment for Electric Ride-Hailing Fleet Control}
\author{
    Tommaso Schettini\textsuperscript{1} \quad
    Nicholas D. Kullman\textsuperscript{2} \quad
    Jorge E. Mendoza\textsuperscript{3} \\[6pt]
    \textsuperscript{1}Concordia University \\
    \texttt{tommaso.schettini@concordia.ca} \\[4pt]
    \textsuperscript{2}Universit\'e de Tours \\
    \texttt{nick.kullman@gmail.com} \\[4pt]
    \textsuperscript{3}HEC Montr\'eal \\
    \texttt{jorge.mendoza@hec.ca}
}
\date{}
\begin{document}
\maketitle
\begin{abstract}
    Machine-learning policies have attracted increasing interest for ride-hailing fleet control in recent years.
    Reinforcement learning, in particular, requires a structured simulation environment that specifies observations, actions, rewards, and decision epochs for training and evaluation.
    For electric fleets, this environment must also capture the interaction among stochastic demand, vehicle operations, and capacitated charging infrastructure.
    We present \sname, an open-source Gymnasium environment for joint control of electric ride-hailing fleets.
    Its fixed-size observation--action interface exposes request assignment, repositioning, and charging to a single policy.
    The event-driven simulator represents requests with pickup deadlines, vehicle job queues, battery dynamics, and finite-capacity charging facilities with first-in--first-out queues.
    A configurable decision-epoch mechanism separates internal simulator events from policy interactions, supporting event-driven, periodic, hybrid, and policy-requested control within the same operational model.
    The software provides seeded instances, feasible-action utilities, evaluation tools, operational metrics, and baseline policies.
    The source code is available at \url{https://github.com/tommaso-schettini/openhail}.
\end{abstract}


\section{Introduction}
\label{sec:introduction}

Ride-hailing operators must respond to requests issued at uncertain times and locations while limiting empty travel and energy consumption.
Traditional fleet-control problems focus on assigning vehicles to requests and repositioning idle vehicles in anticipation of future demand \citep{alonso2017demand,lin2018efficient}.
As ride-hailing fleets become electrified, these problems expand to include charging decisions and additional operational constraints \citep{kullman2022dynamic}.
The operator must determine when and where vehicles should charge while accounting for limited driving ranges, charging times, and competition for capacitated charging infrastructure.
These decisions are highly related, since serving a request consumes energy and delays charging.
Sending a vehicle to charge reduces the fleet available to serve current demand.

Reinforcement learning (RL) is emerging as a promising approach to coordinated fleet repositioning and joint assignment, repositioning, and charging \citep{lin2018efficient,kullman2022dynamic,dai2025atomic}.
These methods shift part of the computational effort to offline training.
While training can be costly, the learned policy or value-function approximations support rapid action selection at deployment.
For instance, the Drafter controller of \cite{kullman2022dynamic} selects assignment, repositioning, and charging actions by evaluating trained neural networks online.
This avoids solving an optimization problem during action selection; for comparison, their reoptimization benchmark solves two mixed-integer programs at each scheduled decision epoch.

Training and evaluating RL policies requires a structured simulation environment that defines observations, actions, rewards, and decision epochs.
Standard interfaces such as Gymnasium separate the implementation of these interactions from the learning algorithm, allowing different policies to use a common environment \citep{towers2024gymnasium}.
In multi-agent settings, the Agent Environment Cycle (AEC) games model makes the sequence of agent actions and environment updates explicit, with agents acting one at a time \citep{terry2021pettingzoo}.
Related efforts include MAEnvs4VRP, which provides modular environments for reinforcement-learning studies of vehicle routing problems \citep{gama2026multiagent}.
An electric-fleet environment must make service and charging decisions explicit and represent their effects on subsequent vehicle availability and energy reserves.
It must also specify when the policy observes the system and acts, since requests, trips, and charging operations evolve asynchronously.

The starting point for \sname\ is \textit{pyhailing}, an open-source OpenAI Gym environment for controlling a homogeneous ride-hailing fleet \citep{pyhail}.
\textit{pyhailing} was designed to support the DIMACS Challenge's dynamic vehicle-routing variant, in which a controller manages a homogeneous fleet serving stochastic requests to maximize daily profit \citep{pyhail}.
This foundation provides the simulation logic for stochastic trip requests, vehicle job queues, request assignment, and repositioning using trip data from Manhattan.

Building on \textit{pyhailing}, we make four contributions:
\begin{enumerate}[label=\arabic*),leftmargin=*]
    \item We provide a Gymnasium interface with fixed-size observations and joint actions for request assignment, repositioning, and charging.
    The interface includes action validation and reusable utilities for constructing feasible-action masks.
    \item We separate internal simulator events from policy decision epochs, supporting event-driven, periodic, hybrid, and policy-requested interaction without changing the operational model.
    \item We add battery dynamics and finite-capacity charging facilities with explicit travel and charging states and first-in--first-out waiting queues.
    \item We provide seeded configurations, baseline policies, evaluation utilities, and automated tests.
    We use these tools to check operational invariants and measure computational scaling under three control workloads.
\end{enumerate}

The remainder of this paper is structured as follows.
Section~\ref{sec:background} positions \sname\ relative to existing simulation environments.
Section~\ref{sec:architecture} presents the architecture, interface, and decision-epoch mechanism of the software.
Section~\ref{sec:experiments} describes the verification and computational-performance experiments.
Section~\ref{sec:conclusion} summarizes the contribution and identifies directions for further research.

\section{Background and Related Software}
\label{sec:background}

\subsection{Simulation Environments for Fleet-Control Research}
\label{sec:background:software}

Research on fleet control has produced a range of open-source tools for simulating transportation services and evaluating operational policies.
We distinguish three overlapping groups of open-source simulation software: transportation-system testbeds, fleet-service simulators, and controller-facing environments.
The first group represents interactions among travelers, vehicles, and the surrounding network; the second concentrates on request and vehicle operations; the third organizes the simulation around a repeated observation--action interface for policy development.

Transportation-system testbeds include AMoDeus and MaaSSim.
\cite{ruch2018amodeus} introduce AMoDeus as an extension of MATSim for autonomous mobility-on-demand services, combining dynamic demand, dispatching algorithms, service-level analysis, and a graphical viewer with an agent-based transport representation.
MaaSSim models the interactions within two-sided mobility platforms \citep{kucharski2022maassim}.
Travelers, drivers, and the platform act as separate decision makers whose behavior can be specified through user-defined Python modules.

Fleet-service simulators focus more directly on the operational evolution of on-demand fleets.
RidePy provides a modular event-based simulator for ride-hailing and ride-pooling, with replaceable request generators, dispatchers, transport spaces, and vehicle representations, as well as performance-critical components in Cython and C++ \citep{jung2024ridepy}.
FleetPy combines assignment, repositioning, routing, and charging modules with support for multiple operators, standardized data sets, and performance indicators \citep{engelhardt2026fleetpy}.
It provides base classes and interfaces for implementing and testing fleet-control algorithms under different service configurations, including immediate- and batch-offer flows.
The simulator proposed by \cite{zhang2024evsim}, which we refer to as EV-Sim, uses SimPy to represent asynchronous passenger matching, vehicle movements, and charging for electric ride-hailing fleets.
It provides hooks for matching and charging algorithms and demonstrates them using New York City taxi data.

Controller-facing environments expose observations and actions through an interface for policy development.
\textit{pyhailing} exposes request assignment and repositioning through OpenAI Gym \citep{pyhail}.
RideGym provides a Gym-like environment and reproducible benchmark for large-scale ride-pooling and order dispatching, with road-network routing, passenger capacity, and impatient customers \citep{zhao2026ridegym}.
FleetPy also provides a Gymnasium wrapper with fixed observation and action spaces for zonal dispatching \citep{fleetpyGym}.

\textit{pyhailing} supplies the principal non-electric operations on which \sname\ builds, but does not represent batteries or charging.
Its observation and action dimensions also depend on the number of pending requests, whereas \sname\ uses fixed-size spaces for compatibility with learning methods that require them.
RideGym emphasizes pooling and dispatching without electric-vehicle or charging decisions; \sname\ instead exposes assignment, repositioning, and charging for an electric ride-hailing fleet.
EV-Sim is close in operational scope, but its algorithm hooks are organized around simulation processes and experiment scripts rather than a standardized joint observation--action interface.
Among the systems reviewed, FleetPy is the closest comparison in terms of electric-fleet operations and Gymnasium support.
Like \sname, it supports assignment, repositioning, and charging through fleet-control modules.
However, its Gymnasium wrapper exposes only zonal dispatching.
By comparison, \sname\ exposes joint request assignment, repositioning, and charging through a single Gymnasium interface with configurable decision epochs.

Table~\ref{tab:related-software} compares service scope, policy interfaces, control decisions, and electric operations.
The control decisions refer to the listed interface, which may expose only part of the capabilities of the wider simulator.

\begin{table}[tb]
    \caption{Service scope and policy interfaces of open-source fleet-control software.}
    \label{tab:related-software}
    \centering
    \footnotesize
    \begin{adjustbox}{max width=\textwidth}
        \begin{tabular}{@{}
                >{\raggedright\arraybackslash}p{1.6cm}
                >{\raggedright\arraybackslash}p{2.2cm}
                >{\raggedright\arraybackslash}p{2.8cm}
                >{\raggedright\arraybackslash}p{3.0cm}
                >{\raggedright\arraybackslash}p{2.4cm}@{}}
            \toprule
            Software & Service scope & Policy interface & Control decisions & Electric operations \\
            \midrule
            AMoDeus & Autonomous mobility & MATSim modules & Dispatching & NR \\
            MaaSSim & Two-sided platforms & Behavioral modules & Traveler, driver, platform & NR \\
            RidePy & Hailing and pooling & Dispatcher modules & Dispatching & NR \\
            FleetPy & Mobility-on-demand & Fleet-control modules & Assignment, repositioning, charging & Battery, charging \\
             & & Gymnasium wrapper$^{a}$ & Zonal dispatching & \\
            EV-Sim & Electric ride-hailing & Algorithm hooks & Matching, charging & Battery, charging \\
            \textit{pyhailing} & Ride-hailing & OpenAI Gym & Assignment, repositioning & None \\
            RideGym & Ride-pooling & Gym-like & Pooling, dispatching & None \\
            \sname & Electric ride-hailing & Gymnasium & Assignment, repositioning, charging, decision timing & Battery, charging, queues \\
            \bottomrule
        \end{tabular}
    \end{adjustbox}
    \par\smallskip
    \begin{minipage}{\textwidth}
        \footnotesize
        NR: not reported as a primary component in the cited source.
        $^{a}$The FleetPy wrapper exposes zonal dispatching; the electric operations listed above belong to the wider simulator.
    \end{minipage}
\end{table}

\section{The \sname\ Architecture}
\label{sec:architecture}

\begin{figure}[tbp]
    \centering
    \includegraphics[width=\textwidth]{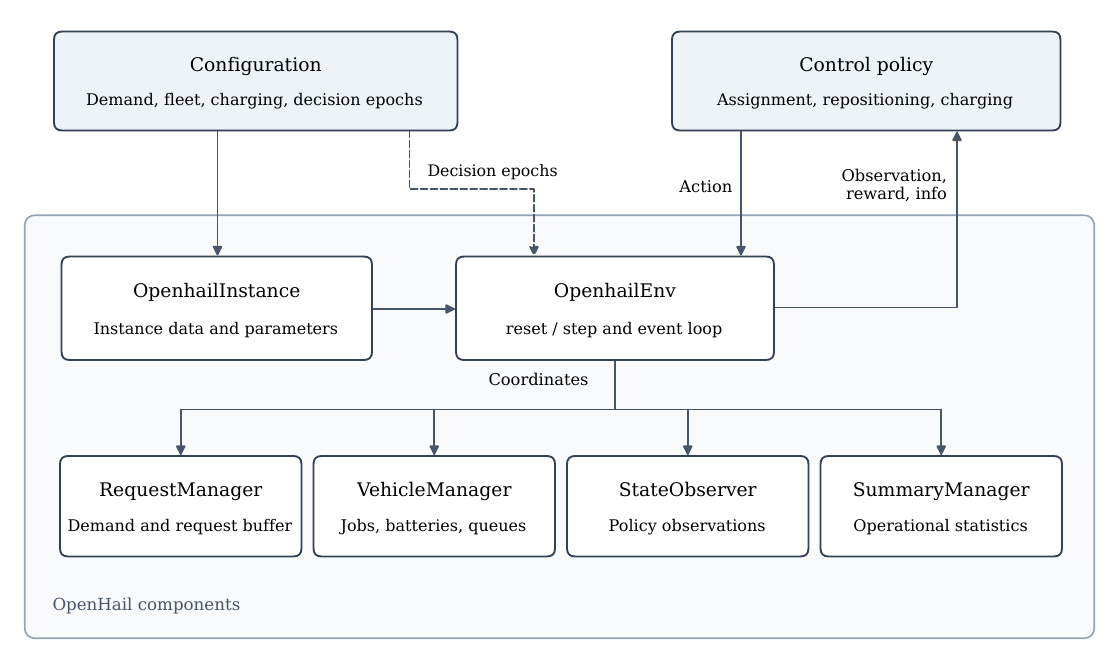}
    \caption{The \sname\ architecture.
    Configuration supplies instance data and decision-epoch settings; policies interact with \texttt{OpenhailEnv}, which coordinates the five supporting components.}
    \label{fig:architecture}
\end{figure}

The architecture comprises six components (Figure~\ref{fig:architecture}):
\begin{itemize}[leftmargin=*]
    \item \texttt{OpenhailInstance} defines the geographic, demand, fleet, charging, and economic inputs.
    \item \texttt{RequestManager} generates the seeded request sequence and maintains the pending-request buffer.
    \item \texttt{VehicleManager} advances vehicle jobs, battery states, and charging operations.
    \item \texttt{StateObserver} constructs observations and checks their consistency with the declared space.
    \item \texttt{SummaryManager} records request, vehicle, charging, and decision-epoch statistics.
    \item \texttt{OpenhailEnv} coordinates these components through the Gymnasium \texttt{reset} and \texttt{step} methods.
\end{itemize}

\subsection{Core Components and Event Dynamics}
\label{sec:architecture:components}

\texttt{OpenhailInstance} combines three forms of input.
Geographic files define service zones and candidate reposition locations; historical trip records provide request times, origins, destinations, and trip distances; and configuration files specify fleet, charging, service, and reward parameters.
The current implementation includes a request generator for New York City.
Configurable parameters include the planning horizon, fleet size, battery capacity, travel speed, discharge and charging rates, passenger pickup limit, and charging capacities.

For an environment constructed from these inputs, we denote by $N$ the number of vehicles, $M$ the maximum number of pending requests exposed to the policy, and $D$ the number of reposition locations available for repositioning or charging.
These dimensions remain fixed during an episode.

\texttt{RequestManager} uses the instance's request generator to construct a seeded request sequence and maintains the fixed-size buffer exposed to the policy.
It releases requests as simulation time advances, removes served requests, and manages unserved requests according to the configured retention and pickup limits.
When new arrivals exceed the buffer capacity, older pending requests are discarded to make room.
Empty positions contain dummy requests to ensure that the observation space remains fixed-size and compatible with learning methods that require fixed input dimensions.
Request generation and vehicle initialization use separate random-number streams derived from the episode seed, allowing the demand realization to be held fixed across policies and decision-epoch configurations.
The full request sequence is generated at the start of each episode, but observations expose only requests released by the current simulation time; future requests remain hidden from the policy.

\texttt{VehicleManager} represents each vehicle through its current location, battery level, job type, scheduled destination, expected completion time, and battery level at completion.
The supported job types are idle, passenger service, repositioning, travel to a charger, waiting for a charger, and charging.
Travel times are obtained from spatial distances and a configured constant speed, while energy consumption and charging duration follow the configured linear discharge and charging rates.
Network congestion and route choice are not modeled.

A policy may send a vehicle to a reposition location without charging or may request charging upon arrival.
If a charging post is available, the vehicle starts charging immediately; otherwise, it enters the station queue.
Available posts are assigned to queued vehicles in first-in--first-out order.
The policy can remove a vehicle from a charging queue by issuing a feasible repositioning action without charging or redirect it to another station by requesting charging there.

\texttt{StateObserver} constructs the policy observation from the request buffer, vehicle states, charging-facility states, simulation time, and current epoch type.
It defines the observation space and provides consistency checks without advancing the simulated system.
The returned vehicle arrays are copies of the internal state.
Section~\ref{sec:architecture:interface} details their structure and dimensions.

\texttt{SummaryManager} accumulates request-service counts, vehicle activity times, charging statistics, and decision-epoch metrics when the corresponding tracking options are enabled.

Requests arrive at customer-specific times, and vehicles complete trips, repositioning movements, and charging operations after different durations.
Event-driven decision models preserve this temporal order and allow a controller to respond when information or vehicle availability changes \citep{menda2019eventdriven}.
However, some events must be processed to update the simulated system even when they do not warrant a new policy decision.
We therefore distinguish between \textit{operational events}, which update the system state, and \textit{decision epochs}, at which the environment returns an observation and requests an action.

In \textit{pyhailing}, new requests and selected service completions trigger decisions, and a maximum interdecision time can prevent long intervals without policy interaction.
\sname\ generalizes this mechanism through an exposure configuration that determines which operational events also produce decision epochs.
Request arrivals, service completions, and charging completions can independently trigger decisions; repositioning completions remain internal state transitions.
A periodic decision clock can operate alone or alongside these event triggers, while a maximum interdecision time limits the interval between policy interactions.
A policy can also request an absolute future decision time in its action.
When request epochs are disabled, arriving requests accumulate until the next exposed epoch, subject to the configured buffer and pickup limits.

\texttt{OpenhailEnv} initializes these components, applies policy actions, and advances the event process until the next exposed decision epoch.
Each call to \texttt{step} consists of four stages.
First, the environment decomposes the joint action and applies feasible service, repositioning, and charging instructions.
Second, it allocates available charging posts and schedules the resulting vehicle jobs.
Third, it advances to the earliest eligible time among the planning horizon, the next vehicle-job completion, a periodic or maximum-interdecision clock, a policy-requested time, and, when enabled, the next request arrival.
Vehicle and charging events encountered before a decision epoch are processed internally, after which the event search continues.
Fourth, the environment releases all requests issued by the selected time, constructs the new observation, and returns control to the policy.

\subsection{Library Implementation and Interface}
\label{sec:architecture:interface}

The controller-facing interface follows the Gymnasium application programming interface.
Calling \texttt{reset(seed=s)} initializes the demand and vehicle substreams and returns an observation, while \texttt{step(action)} returns the next observation, the reward accumulated since the preceding decision epoch, termination and truncation indicators, and an information dictionary.
The environment terminates at the configured planning horizon.

Table~\ref{tab:observation-space} describes the five elements of the observation.
Vehicle arrays contain one row per fleet vehicle, request arrays contain $M$ rows, and charger arrays contain one row per reposition location.
When fewer than $M$ requests are pending, the unused rows contain designated dummy requests whose issue times fall beyond the episode horizon.

\begin{table}[tb]
    \caption{Elements of the \sname\ observation space.}
    \label{tab:observation-space}
    \centering
    \footnotesize
    \begin{adjustbox}{max width=\textwidth}
        \begin{tabular}{@{}
                >{\raggedright\arraybackslash}p{2.6cm}
                >{\raggedright\arraybackslash}p{3.2cm}
                >{\raggedright\arraybackslash}p{9.3cm}@{}}
            \toprule
            Entry & Shape & Information \\
            \midrule
            \texttt{request} & $M$ request records & Issue time, origin, destination, and trip distance for each pending request. \\
            \texttt{V} & $N$ vehicle records & Current and scheduled locations, times, and battery levels; job type; current and target charger; and next vehicle-event time. \\
            \texttt{chargers} & $D$ facility records & Reposition coordinates, charging capacity, current occupancy, and queue length. \\
            \texttt{time} & Scalar & Current simulation time. \\
            \texttt{epoch\_type} & Scalar & Event type that produced the current decision epoch. \\
            \bottomrule
        \end{tabular}
    \end{adjustbox}
\end{table}

The action is a tuple with three elements, as reported in Table~\ref{tab:action-space}.
The service vector assigns at most one vehicle identifier to each pending-request position, with $N$ serving as the sentinel for leaving a request unassigned.
The repositioning vector contains one entry per vehicle.
For a zero-indexed reposition location $d$, action $2d+1$ sends the vehicle to $d$ without requesting charge, while action $2d+2$ sends it to the same location and requests charge.
Finally, action zero leaves the current movement or charging intention unchanged.
The final scalar requests the absolute time of a future decision epoch.
A nonpositive value requests no additional epoch.

\begin{table}[tb]
    \caption{Elements of the \sname\ action space.}
    \label{tab:action-space}
    \centering
    \footnotesize
    \begin{adjustbox}{max width=\textwidth}
        \begin{tabular}{@{}
                >{\raggedright\arraybackslash}p{2.8cm}
                >{\raggedright\arraybackslash}p{3.1cm}
                >{\raggedright\arraybackslash}p{9.2cm}@{}}
            \toprule
            Entry & Gymnasium space & Interpretation \\
            \midrule
            \texttt{serve} & \texttt{MultiDiscrete}, length $M$ & Vehicle assigned to each pending request, or the unassigned-request sentinel. \\
            \texttt{reposition} & \texttt{MultiDiscrete}, length $N$ & No-op, reposition-only, or reposition-and-charge instruction for each vehicle. \\
            \texttt{epoch} & Scalar \texttt{Box} & Absolute future time at which the policy requests another decision opportunity. \\
            \bottomrule
        \end{tabular}
    \end{adjustbox}
\end{table}

A service assignment must satisfy the passenger pickup limit and the vehicle's energy requirement.
A repositioning instruction must be reachable with the available battery level.

For convenience, \sname\ provides functions that compute assignment and repositioning masks from an observation, allowing a policy to filter infeasible choices before selecting an action.
Strict-validation mode additionally checks service and repositioning feasibility as actions are processed by \texttt{step}.
A detected violation raises an exception and interrupts the rollout.

The scalar reward represents net operating value over the interval between decisions.
Serving a request generates revenue with a fixed component and a distance-dependent component, from which pickup and passenger-carrying travel costs are deducted.
Repositioning travel and charging energy generate additional costs as the event process advances.
The accompanying information dictionary separates service and operating reward components and reports the elapsed simulation time, triggering epoch type, numbers of added and served requests, mean state of charge, and charger occupancy and queue lengths.

The configuration runner automates experiments across policies, infrastructure alternatives, and random seeds.
Researchers can compare decision schedules under common demand realizations to examine how the frequency of policy decisions affects computational workload and service outcomes.
They can also vary charging capacity to study how policies allocate vehicles between passenger service, repositioning, and charging when vehicles compete for available posts.

\section{Verification and Computational Performance}
\label{sec:experiments}

We measure wall-clock runtime per episode and per decision epoch as fleet size and the number of reposition locations increase, with request volume proportional to fleet size.

\subsection{Software Verification}
\label{sec:experiments:verification}

Unit and integration tests cover seeded demand, Gymnasium observations and actions, time bounds, energy and reward accounting, and charging queues, including simultaneous charging completions.
Additional tests cover complete baseline-policy episodes, rendering without state mutation, and PPO checkpoint loading and saving.
Test descriptions and execution commands are provided in the package documentation.

\subsection{Experimental Setup}
\label{sec:experiments:performance}

\subsubsection{One-Day Policy Evaluation}
\label{sec:experiments:evaluation}

Each evaluation episode represents one day of electric ride-hailing operations.
We vary fleet size $N\in\{100,500,1{,}000,2{,}000\}$ and the number of reposition locations $D\in\{5,10,20,40\}$, yielding 16 configurations.
Each instance contains $20N$ requests and $0.2N$ charging posts.
Approximately one quarter of the reposition locations host charging facilities, and the posts are distributed uniformly among these locations.
This construction holds demand per vehicle and charging capacity per vehicle constant while varying fleet size and the number of available destinations.

The software provides nearest and random-feasible baseline agents and a periodic proximal policy optimization (PPO) agent that can load pretrained checkpoints.
The \emph{nearest} policy uses distance-based assignment and repositioning rules.
The \emph{random-feasible} policy enumerates feasible repositioning and charging actions before sampling an action at 15-minute clock epochs.
The \emph{periodic PPO} policy uses the same clock epochs but selects repositioning and charging actions through neural-network inference.
For the learned-policy workload, we use a separate pretrained checkpoint for each configuration, obtained as described in Section~\ref{sec:experiments:training}.
Both periodic controllers use nearest-feasible assignment at request epochs; their periodic decisions concern repositioning and charging.
At other non-clock epochs, they issue no repositioning or charging action unless a jobless vehicle requires a nearest-feasible repositioning instruction.

We evaluate each controller on five paired one-day instances for every $(N,D)$ configuration, yielding 240 measured rollouts.
PPO uses sampled actions from the selected checkpoint without parameter updates.
The controllers share environment seeds within each configuration; stochastic controllers also use matched policy seeds.
We perform one unmeasured warm-up rollout per controller and configuration and rotate controller execution order across replications.
Measurements are collected sequentially within a single Compute Canada allocation on an AMD EPYC 7532 processor, with the process bound to one CPU.
The environment uses Linux, Python 3.13.2, NumPy 2.4.2, and PyTorch 2.13.0; numerical-library and PyTorch thread counts are fixed at one.
Rendering, strict validation, and activity tracking are disabled.

The timed interval begins after instance reset and controller initialization and ends at episode termination.
We record total wall-clock duration, accumulated time in controller calls, and accumulated time in environment transitions; total duration additionally includes measurement-loop overhead.
Dividing each rollout duration by its number of exposed decision epochs gives the time per epoch.
We report medians and interquartile ranges across the five paired instances for each controller and configuration.

\begin{figure}[t]
    \centering
    \includegraphics[width=\textwidth]{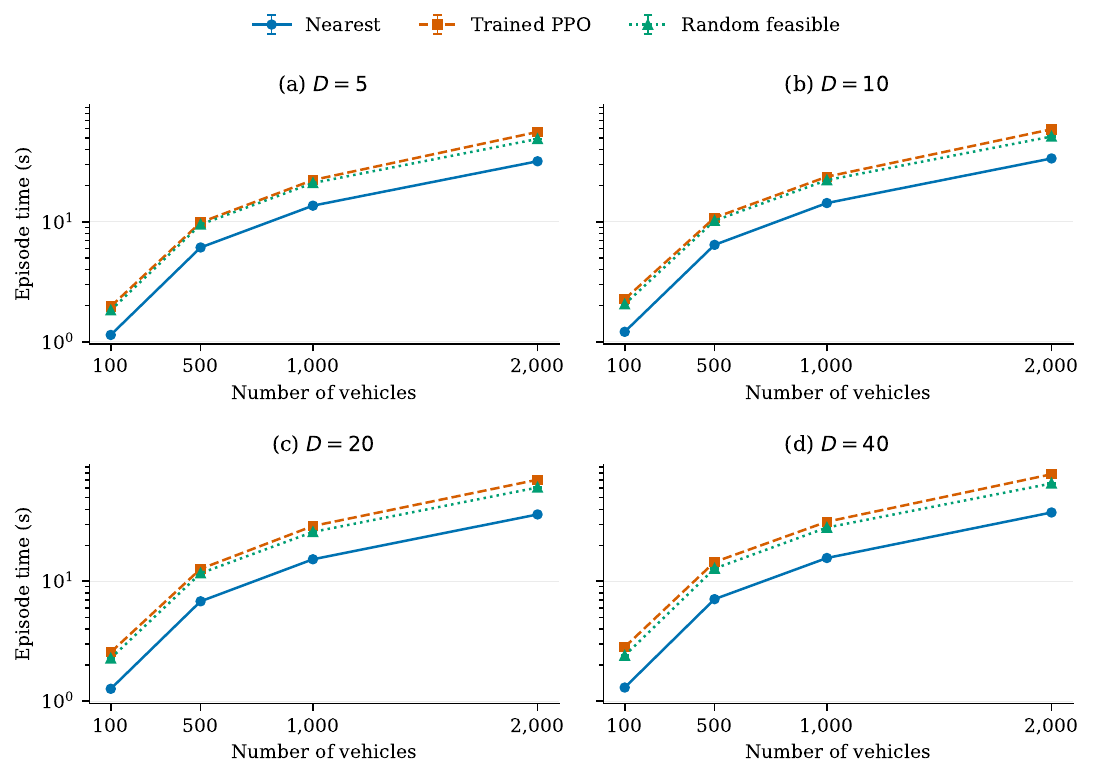}
    \caption{Post-hoc simulation time by controller, fleet size, and location count.
    Each marker is the median duration of five paired one-day rollouts; error bars span the interquartile range.
    The vertical axes are logarithmic.
    Policy optimization, model loading, and reset are excluded.}
    \label{fig:posthoc-episode-times}
\end{figure}

\subsubsection{PPO Training}
\label{sec:experiments:training}

For each configuration, we train a separate periodic PPO policy for 500 episodes.
Policy evaluation occurs every 25 episodes on five fixed environment--policy seed pairs, and the checkpoint with the highest mean evaluation reward is retained for the one-day experiments.
Training uses CPU execution with validation and activity tracking enabled.
Table~\ref{tab:training-cost} reports the elapsed time to train each agent, including periodic evaluation.
Training requires 0.54--21.03 hours per configuration.
At each location count, larger fleets require more time; the effect of increasing the number of locations is not monotone across the tested configurations.

\begin{table}[tbp]
    \centering
    \caption{Elapsed time (hours) to train each PPO agent for 500 episodes, including periodic evaluation.
    Columns give the number of reposition locations.}
    \label{tab:training-cost}
    \begin{tabular}{rrrrr}
        \toprule
        & \multicolumn{4}{c}{Reposition locations} \\
        \cmidrule(lr){2-5}
        Vehicles & 5 & 10 & 20 & 40 \\
        \midrule
        100 & 0.54 & 0.60 & 0.82 & 0.75 \\
        500 & 2.74 & 2.85 & 3.27 & 10.44 \\
        1,000 & 6.56 & 6.56 & 13.85 & 8.44 \\
        2,000 & 16.29 & 16.55 & 18.76 & 21.03 \\
        \bottomrule
    \end{tabular}
\end{table}

\begin{figure}[t]
    \centering
    \includegraphics[width=\textwidth]{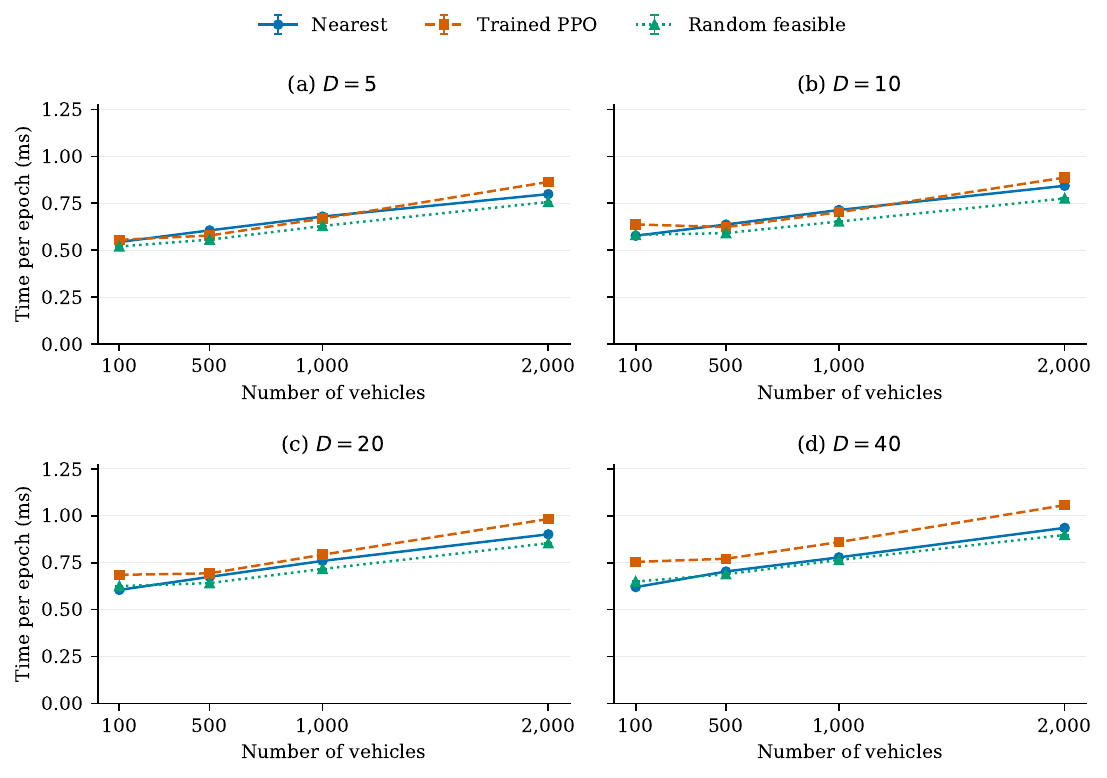}
    \caption{Post-hoc simulation time per exposed decision epoch.
    Each marker is the median of the five rollout-specific ratios of episode duration to decision count; error bars span the interquartile range.
    All panels use the same linear vertical scale.}
    \label{fig:posthoc-epoch-times}
\end{figure}

\subsection{Post-hoc Simulation Performance}
\label{sec:experiments:baselines}

Figure~\ref{fig:posthoc-episode-times} compares episode duration across controllers.
Each panel fixes the location count, with fleet size on the horizontal axis and wall-clock time in seconds on the logarithmic vertical axis.
Figure~\ref{fig:posthoc-epoch-times} uses the same arrangement to report time per exposed decision epoch in milliseconds.

Median episode duration increases with fleet size for all three controllers.
At 100 vehicles, it ranges from 1.14--1.30 seconds for nearest, 1.83--2.40 seconds for random feasible, and 1.97--2.83 seconds for PPO across the four location counts.
At 2,000 vehicles, the corresponding ranges are 32.03--37.53, 49.15--65.51, and 56.25--78.22 seconds.
PPO requires 1.62--2.18 times the episode duration of nearest over the tested configurations.
At 2,000 vehicles, median time per epoch is 0.799--0.936 milliseconds for nearest, 0.757--0.898 milliseconds for random feasible, and 0.864--1.057 milliseconds for PPO.

Table~\ref{tab:posthoc-decomposition} decomposes the runtime of the largest configuration, with 2,000 vehicles and 40 reposition locations.
Across all tested configurations, median controller computation time is lower than median environment-update time for all three policies.
PPO exposes approximately 84\% more decision epochs than nearest, with a median time per epoch that is approximately 13\% higher.
Relative to random feasible, the additional PPO runtime is concentrated in controller computation: median agent time is 27.36 versus 15.67 seconds, whereas environment time is 50.85 versus 49.78 seconds.

\begin{table}[tbp]
    \centering
    \caption{Post-hoc runtime decomposition for 2,000 vehicles and 40 reposition locations.
    Each entry is the median over five paired rollouts; wall time also includes measurement-loop overhead.}
    \label{tab:posthoc-decomposition}
    \begin{adjustbox}{max width=\textwidth}
        \begin{tabular}{lrrrrr}
        \toprule
        Controller & Wall (s) & Agent (s) & Environment (s) & Epochs & Time/epoch (ms) \\
        \midrule
        Nearest & 37.53 & 15.68 & 21.83 & 40,096 & 0.936 \\
        Random feasible & 65.51 & 15.67 & 49.78 & 72,666 & 0.898 \\
        Trained PPO & 78.22 & 27.36 & 50.85 & 73,929 & 1.057 \\
        \bottomrule
        \end{tabular}
    \end{adjustbox}
\end{table}

\section{Conclusion}
\label{sec:conclusion}

In this paper, we introduce \sname, an open-source environment for developing and evaluating control policies for electric ride-hailing fleets.
The library provides a common Gymnasium interface for request assignment, repositioning, and charging, allowing policies to account for the interactions among passenger service, vehicle availability, and charging queues.

\sname\ has been designed using a modular architecture that allows researchers to adapt the environment to different fleet configurations and control approaches.
The New York City request generator, configurable charging infrastructure, and baseline policies provide initial settings from which to develop such studies.
Alongside the library, we provide training and evaluation tools that illustrate how reinforcement-learning policies can be integrated into the environment and evaluated alongside other controllers.
In the largest tested configuration, with 2,000 vehicles and 40 reposition locations, median wall-clock runtime per simulated day ranges from 37.53 to 78.22 seconds across the three controllers on one allocated CPU.

\sname\ also provides a mechanism for controlling the frequency and timing of policy decisions and selecting the types of operational events that trigger them.
Further research can examine how decision timing affects the quality of fleet control by comparing policies that respond to operational events, act periodically, or request their next decision time.
With these mechanisms and its modular architecture, \sname\ aims to serve as both a practical tool for developing and evaluating fleet-control policies and a foundation for future research on electric ride-hailing operations.

\section*{Software Availability}

\sname\ is available under the MIT license at \url{https://github.com/tommaso-schettini/openhail}, together with documentation and reproduction instructions.

\bibliography{Wrath}

@article{gama2026multiagent,
  title={Multiagent Environments for Vehicle Routing Problems},
  author={Gama, Ricardo and Cunha, Ricardo and Fuertes, Daniel and del-Blanco, Carlos R. and Fernandes, Hugo L.},
  journal={INFORMS Journal on Computing},
  year={2026},
  note={Articles in Advance},
  doi={10.1287/ijoc.2025.1211}
}

@inproceedings{lin2018efficient,
  title={Efficient Large-Scale Fleet Management via Multi-Agent Deep Reinforcement Learning},
  author={Lin, Kaixiang and Zhao, Renyu and Xu, Zhe and Zhou, Jiayu},
  booktitle={Proceedings of the 24th ACM SIGKDD International Conference on Knowledge Discovery and Data Mining},
  year={2018},
  url={https://www.kdd.org/kdd2018/accepted-papers/view/efficient-large-scale-fleet-management-via-multi-agent-deep-reinforcement-l}
}

@article{kullman2022dynamic,
  title={Dynamic ride-hailing with electric vehicles},
  author={Kullman, Nicholas D and Cousineau, Martin and Goodson, Justin C and Mendoza, Jorge E},
  journal={Transportation Science},
  volume={56},
  number={3},
  pages={775--794},
  year={2022},
  publisher={INFORMS}
}

@article{alonso2017demand,
  title={On-demand high-capacity ride-sharing via dynamic trip-vehicle assignment},
  author={Alonso-Mora, Javier and Samaranayake, Samitha and Wallar, Alex and Frazzoli, Emilio and Rus, Daniela},
  journal={Proceedings of the National Academy of Sciences},
  volume={114},
  number={3},
  pages={462--467},
  year={2017},
  publisher={National Acad Sciences}
}

@misc{pyhail,
  title = {{pyhailing}},
  author = {Kullman, Nicholas},
  year = {2022},
  version = {0.0.9},
  url = {https://pypi.org/project/pyhailing/},
  note = {Python package}
}

@inproceedings{ruch2018amodeus,
  title = {{AMoDeus}, a simulation-based testbed for autonomous mobility-on-demand systems},
  author = {Ruch, Claudio and H\"orl, Sebastian and Frazzoli, Emilio},
  booktitle = {2018 21st International Conference on Intelligent Transportation Systems},
  pages = {3639--3644},
  year = {2018},
  organization = {IEEE},
  doi = {10.1109/ITSC.2018.8569961}
}

@article{kucharski2022maassim,
  title = {Simulating two-sided mobility platforms with {MaaSSim}},
  author = {Kucharski, Rafa{\l} and Cats, Oded},
  journal = {PLOS ONE},
  volume = {17},
  number = {6},
  pages = {e0269682},
  year = {2022},
  doi = {10.1371/journal.pone.0269682}
}

@article{jung2024ridepy,
  title = {{RidePy}: A fast and modular framework for simulating ridepooling systems},
  author = {Jung, Felix and Manik, Debsankha},
  journal = {Journal of Open Source Software},
  volume = {9},
  number = {97},
  pages = {6241},
  year = {2024},
  doi = {10.21105/joss.06241}
}

@article{engelhardt2026fleetpy,
  title = {{FleetPy}: An open source simulator for reproducible research on mobility-on-demand services},
  author = {Engelhardt, Roman and Dandl, Florian and Syed, Arslan Ali and Ding, Chenhao and Alvarez-Ossorio Martinez, Santiago and Zhang, Yunfei and Hamdy, Hoda and Brodersen, Joel and Chen, Zhipu and Bogenberger, Klaus},
  journal = {European Transport Research Review},
  volume = {18},
  pages = {52},
  year = {2026},
  doi = {10.1186/s12544-026-00823-3}
}

@misc{zhang2024evsim,
  title = {A Simulation Framework for Ride-Hailing with Electric Vehicles},
  author = {Zhang, Chen and Varma, Sushil},
  year = {2024},
  eprint = {2411.19471},
  archivePrefix = {arXiv},
  primaryClass = {math.OC},
  doi = {10.48550/arXiv.2411.19471}
}

@misc{zhao2026ridegym,
  title = {{RideGym}: A Standardized Interface for Real-World Large-Scale Ride-Sharing Systems},
  author = {Zhao, Zijian and Hu, Yulong and Li, Sen},
  year = {2026},
  eprint = {2607.10173},
  archivePrefix = {arXiv},
  primaryClass = {cs.MA},
  url = {https://arxiv.org/abs/2607.10173}
}

@article{menda2019eventdriven,
  title = {Deep Reinforcement Learning for Event-Driven Multi-Agent Decision Processes},
  author = {Menda, Kunal and Chen, Yi-Chun and Grana, Justin and Bono, James W. and Tracey, Brendan D. and Kochenderfer, Mykel J. and Wolpert, David},
  journal = {IEEE Transactions on Intelligent Transportation Systems},
  volume = {20},
  number = {4},
  pages = {1259--1268},
  year = {2019},
  doi = {10.1109/TITS.2018.2868268}
}

@misc{towers2024gymnasium,
      title={Gymnasium: A Standard Interface for Reinforcement Learning Environments},
      author={Mark Towers and Ariel Kwiatkowski and Jordan Terry and John U. Balis and Gianluca De Cola and Tristan Deleu and Manuel Goulão and Andreas Kallinteris and Markus Krimmel and Arjun KG and Rodrigo Perez-Vicente and Andrea Pierré and Sander Schulhoff and Jun Jet Tai and Hannah Tan and Omar G. Younis},
      year={2024},
      eprint={2407.17032},
      archivePrefix={arXiv},
      primaryClass={cs.LG},
      url={https://arxiv.org/abs/2407.17032},
}

@misc{fleetpyGym,
  author={{FleetPy Developers}},
  title={{FleetPy}: {Gymnasium} Environment for Zonal Dispatching},
  year={2026},
  howpublished={Source code, \texttt{FleetPy\_gym.py}},
  url={https://github.com/TUM-VT/FleetPy/blob/main/FleetPy_gym.py},
  note={Accessed September 9, 2026}
}

@inproceedings{terry2021pettingzoo,
  title = {{PettingZoo}: Gym for Multi-Agent Reinforcement Learning},
  author = {Terry, J and Black, Benjamin and Grammel, Nathaniel and Jayakumar, Mario and Hari, Ananth and Sullivan, Ryan and Santos, Luis S and Dieffendahl, Clemens and Horsch, Caroline and Perez-Vicente, Rodrigo and Williams, Niall and Lokesh, Yashas and Ravi, Praveen},
  booktitle = {Advances in Neural Information Processing Systems},
  volume = {34},
  year = {2021},
  url = {https://papers.nips.cc/paper/2021/hash/7ed2d3454c5eea71148b11d0c25104ff-Abstract.html}
}

@misc{dai2025atomic,
  title = {Atomic Proximal Policy Optimization for Electric Robo-Taxi Dispatch and Charger Allocation},
  author = {Dai, Jim and Wu, Manxi and Zhang, Zhanhao},
  year = {2025},
  eprint = {2502.13392},
  archivePrefix = {arXiv},
  primaryClass = {cs.AI},
  doi = {10.48550/arXiv.2502.13392},
  url = {https://arxiv.org/abs/2502.13392}
}

\end{document}